\documentclass[11pt,a4paper]{article}
\usepackage[hyperref]{emnlp2020}
\usepackage[ruled,vlined]{algorithm2e}
\usepackage{times}
\usepackage{latexsym}
\usepackage{amsmath}

\usepackage{graphicx}
\usepackage{amsfonts}
\usepackage{colortbl}
\usepackage{microtype}
\usepackage{multirow, booktabs}

\aclfinalcopy 

\title{RomeBERT: Robust Training of Multi-Exit BERT}

\author{Shijie Geng$^{1}$ \quad Peng Gao$^2$ \quad Zuohui Fu$^{1}$ \quad Yongfeng Zhang$^1$\\
  $^1$Rutgers University \quad $^2$The Chinese University of Hong Kong
}

\date{}

\begin{document}
\maketitle
\begin{abstract}
BERT has achieved superior performances on Natural Language Understanding (NLU) tasks. However, BERT possesses a large number of parameters and demands certain resources to deploy. For acceleration, Dynamic Early Exiting for BERT (DeeBERT) has been proposed recently, which incorporates multiple exits and adopts a dynamic early-exit mechanism to ensure efficient inference. While obtaining an efficiency-performance tradeoff, the performances of early exits in multi-exit BERT are significantly worse than late exits. In this paper, we leverage gradient regularized self-distillation for \textbf{RO}bust training of \textbf{M}ulti-\textbf{E}xit BERT (RomeBERT), which can effectively solve the performance imbalance problem between early and late exits. Moreover, the proposed RomeBERT adopts a one-stage joint training strategy for multi-exits and the BERT backbone while DeeBERT needs two stages that require more training time.
Extensive experiments on GLUE datasets are performed to demonstrate the superiority of our approach. Our code is available at \url{https://github.com/romebert/RomeBERT}.
\end{abstract}

\begin{figure*}
\begin{center}
\includegraphics[width=0.98\linewidth,trim={0cm 0cm 0cm 0cm}]{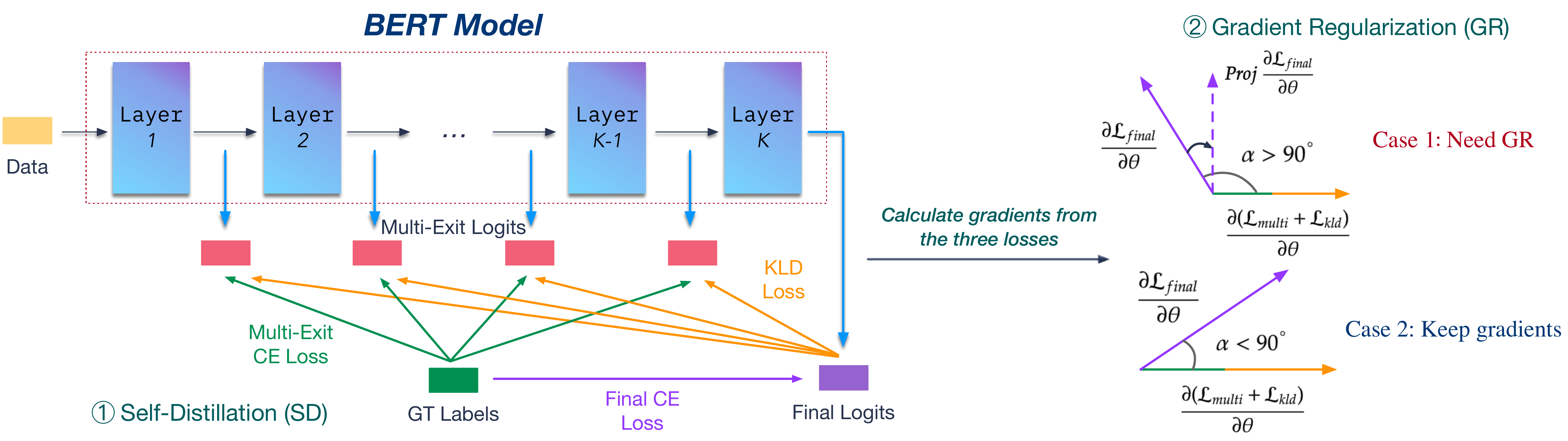}
\end{center}
\caption{RomeBERT includes two key components: 1) Self-Distillation (jointly training multi-exits without freezing the BERT backbone) and 2) Gradient Regularization.}
\label{fig:model}
\end{figure*}

\section{Introduction}
BERT has achieved unprecedented performance on text understanding~\cite{devlin2018bert}. However, BERT is hard to deploy due to its computation cost. Adaptive Neural Networks (ANN)~\cite{huang2016deep,huang2017multi} solve the problem by adding an exit for each block and allows early-exit for simple examples to save computation time. Inspired by ANN, DeeBERT~\citep{Xin2020deebert} proposed multi-exit BERT, which inserts additional classifiers for each layer of the Transformer~\cite{vaswani2017attention}. DeeBERT adopts a two-stage training style: it first fine-tunes BERT on downstream tasks as usual, and then all exits are trained jointly with the main branch of Transformer frozen. However, the dilemma between performance and computation cost still exists. Although DeeBERT demonstrates the feasibility of ANN on BERT, it suffers from the following problems: 1) Significant drop of the performance for early exits, and 2) Freezing the BERT backbone limits the expressive power of BERT for early exits.

To solve the above problems, we propose RomeBERT based on gradient regularized self-distillation. In RomeBERT, we jointly train the weights of BERT backbone with multi-exit classifiers. However, simply unfreezing the BERT backbone will introduce gradient conflicts during training, and thus hurt the performance of both early and late exits. RomeBERT introduces two techniques for robust training of Multi-Exit BERT, namely Gradient Regularization (GR) and Self-Distillation (SD). SD allows early exits to mimic the soft label of the final exit, and thus enables the consistency across different exits. As a result, SD can increase the performance of early exits while keeping the performance of the final exit comparable to the origin model. Besides SD, we adopt a novel GR approach to balance the gradients for early and late exits. 
Extensive ablation studies have been performed on GLUE~\cite{wang2018glue} with the pre-trained BERT model. Experimental results demonstrate that RomeBERT achieves much better balance between efficiency and performance than DeeBERT. 


\section{Related Work}
Transformer has achieved success on both visual and textual understanding~\cite{devlin2018bert,wang2018non,tan2019lxmert,geng2020character,geng2020spatio}. However, Transformer has fixed structure and is hard to deploy in practice due to the high computation cost. Many variants of Transformer have been proposed for improving the computation efficiency by means of feature clustering~\cite{vyas2020fast,zheng2020end,kitaev2020reformer}, low rank approximation~\cite{choromanski2020rethinking}, attention sparse regularization~\cite{gao2021fast}, and reversible architecture~\cite{kitaev2020reformer}. Among these variants, we focus on improving Transformer using adaptive architecture. Depth-adaptive neural networks~\cite{figurnov2017spatially,graves2016adaptive,gao2020multi,Xin2020deebert,zhou2020bert,liu2020fastbert} can perform adaptive inference by only activating part of the network thus can reduce computation cost of Transformer. During training, multi-exit classifiers with various depths are trained together. DeeBERT~\cite{Xin2020deebert} tests the feasibility of depth-adaptive inference on multi-exit BERT by exiting early if the confidence criterion is met. PABEE~\cite{zhou2020bert} takes into consideration the agreement among multi-exits and explores inference with multiple classifiers. 

Knowledge Distillation (KD) \cite{hinton2015distilling} is a technique for network compression. It has been widely applied in CNN based models~\cite{furlanello2018born,xie2019self,yang2019training}. Recently, DistilBERT \cite{sanh2019distilbert} applies knowledge distillation on the BERT model.  Different from knowledge distillation across models in DistilBERT, self-distillation transfer knowledge from late exits to early exits within a model. FastBERT~\cite{liu2020fastbert} proposes a similar structure as DeeBERT by adding knowledge transfer from deep exits to shallow exits. However, knowledge transfer across multi-exits is weak in FastBERT because the whole BERT backbone has been frozen for training stability. In contrast, our RomeBERT jointly trains the BERT backbone and multi-exit classifiers and adds a novel gradient regularization to solve the training stability problem. RomeBERT aims at improving earlier exit classifiers to reduce the performance imbalance among multi-exits. Our gradient regularization is highly motivated by the Gradient Episodic Memory (GEM)~\cite{lopez2017gradient} for life-long learning and Gradient Surgery~\cite{yu2020gradient} for multi-task learning.



\section{Robust Training of Multi-Exit BERT}
\subsection{Revisit DeeBERT}
DeeBERT proposes a two-stage training strategy for fine-tuning the BERT model on downstream NLP tasks. For efficient inference, it inserts an extra classifier (also called \textit{off-ramp}) after each intermediate Transformer layer. Given a data-label pair $(x, y)$ and a pretrained BERT model with $k$ layers. We can denote the output logits of the $i$-th classifier $f_i$ as $f_i(x;\theta_i)$, where $\theta_i$ is the parameters related to $f_i$.
The fine-tuning process of DeeBERT consists of two stages. First, the original BERT components (including BERT embeddings, all Transformer layers, and the final classifier) are fine-tuned with a cross-entropy loss from the final classifier $\mathcal{L}_{ce}(y, f_k(x;\theta_k))$.
Afterwards, all parameters fine-tuned in the first stage are frozen, and all intermediate classifiers but the final one are fine-tuned with the following loss: $\sum^{k-1}_{i=1}\mathcal{L}_{ce}(y, f_{i}(x;\theta_i))$. During inference, DeeBERT can dynamically determine when to early exit according to the confidence of a certain example by setting an entropy threshold $S$. If the entropy of the input example at the $i$-th off-ramp is below $S$ for the first time, the example will exit instantly at layer $i$.

\begin{table*}[htp!]
\scalebox{0.78}{
\begin{tabular}{ccccccccccccc}
\toprule
 & \multicolumn{2}{c}{\textbf{SST-2}} & \multicolumn{2}{c}{\textbf{MRPC}} &  \multicolumn{2}{c}{\textbf{QQP}} & \multicolumn{2}{c}{\textbf{QNLI}}  & \multicolumn{2}{c}{\textbf{MNLI (m/mm)}}  & \multicolumn{2}{c}{\textbf{RTE}}
\\
\cmidrule(lr){2-3}\cmidrule(lr){4-5}\cmidrule(lr){6-7}\cmidrule(lr){8-9}\cmidrule(lr){10-11}\cmidrule(lr){12-13}
Methods & Acc$\%$  & Time$\%$ & $\mathrm{F}_{1}\%$ & Time$\%$ & $\mathrm{F}_{1}\%$ & Time$\%$ &  Acc$\%$ & Time$\%$ & Acc$\%$ & Time$\%$ & Acc$\%$ & Time$\%$ \\
\cmidrule(lr){1-13}
BERT-base  & 93.5 & 100 & 86.9 & 100 & 71.3 & 100 & 90.8 & 100 & 84.1/83.5 & 100 & 69.4 & 100 \\
\cmidrule(lr){1-13}
DeeBERT & 90.5 & 46.6 & 87.4 & 73.9 & 69.7 & 42.6 & 87.1 & 49.3 & 83.4/82.3 & 73.1 & 67.9 & 94.8 \\
DeeBERT+SD & 91.2 & 45.9 & 86.7 & 65.4 & 70.1 & 44.6 & 86.6 & 47.9 & 83.4/82.6 & 73.7 & 67.6 & 90.6 \\
RomeBERT & \cellcolor{cyan}\bf90.6 & \cellcolor{cyan}\bf26.8 & \cellcolor{cyan}\bf86.7 & \cellcolor{cyan}\bf69.1 & \cellcolor{cyan}\bf70.8 & \cellcolor{cyan}\bf34.8 & \cellcolor{cyan}\bf88.7 & \cellcolor{cyan}\bf34.1 & \cellcolor{cyan}\bf83.3/82.6 & \cellcolor{cyan}\bf53.5 & \cellcolor{cyan}\bf69.5 & \cellcolor{cyan}\bf89.4  \\
\cmidrule(lr){1-13}
DeeBERT & 89.0 & 39.2 & 86.4 & 65.7 & 67.8 & 35.3 & 85.5 & 42.4 & 79.2/78.3 & 59.3 & 67.6 & 90.9 \\
DeeBERT+SD & 88.7 & 35.3 & 86.7 & 65.4 & 68.5 & 37.2 & 85.4 & 43.1 & 79.8/78.5 & 59.3 & 67.4 & 86.9 \\
RomeBERT & \cellcolor{cyan}\bf89.0 & \cellcolor{cyan}\bf19.8 & \cellcolor{cyan}\bf86.2 & \cellcolor{cyan}\bf62.7 & \cellcolor{cyan}\bf69.7 & \cellcolor{cyan}\bf22.8 & \cellcolor{cyan}\bf86.1 & \cellcolor{cyan}\bf24.0 & \cellcolor{cyan}\bf82.2/81.4 & \cellcolor{cyan}\bf38.9 & \cellcolor{cyan}\bf68.9 & \cellcolor{cyan}\bf69.2 \\
\cmidrule(lr){1-13}
DeeBERT  & 87.7 & 33.4 & 83.6 & 41.0 & 59.1 & 19.3 & 82.4 & 34.0 & 76.3/75.5 & 53.9 & 66.1 & 79.0 \\
DeeBERT+SD & 88.2 & 32.9 & 84.4 & 40.4 & 59.9 & 20.0 & 81.7 & 34.3 & 76.6/75.4 & 52.8 & 66.3 & 71.0 \\
RomeBERT & \cellcolor{cyan}\bf88.7 & \cellcolor{cyan}\bf15.0 & \cellcolor{cyan}\bf83.8 & \cellcolor{cyan}\bf33.1 & \cellcolor{cyan}\bf67.7 & \cellcolor{cyan}\bf15.2 & \cellcolor{cyan}\bf83.2 & \cellcolor{cyan}\bf19.2 & \cellcolor{cyan}\bf77.6/76.9 & \cellcolor{cyan}\bf21.4 & \cellcolor{cyan}\bf66.9 & \cellcolor{cyan}\bf45.2\\
\bottomrule
\end{tabular}
}
\caption{Performance comparison on the \textsc{test} splits of the GLUE benchmark between DeeBERT and RomeBERT. 
}
\vspace{0.2cm}
\label{tab:main}
\end{table*}

\begin{table*}[t]
\scalebox{0.78}{
\begin{tabular}{ccccccccccccc}
\toprule
 & \multicolumn{2}{c}{\textbf{SST-2}} & \multicolumn{2}{c}{\textbf{MRPC}} &  \multicolumn{2}{c}{\textbf{QQP}} & \multicolumn{2}{c}{\textbf{QNLI}}  & \multicolumn{2}{c}{\textbf{MNLI (m/mm)}}  & \multicolumn{2}{c}{\textbf{RTE}}
\\
\cmidrule(lr){2-3}\cmidrule(lr){4-5}\cmidrule(lr){6-7}\cmidrule(lr){8-9}\cmidrule(lr){10-11}\cmidrule(lr){12-13}
Methods & Acc$\%$  & Time$\%$ & $\mathrm{F}_{1}\%$ & Time$\%$ & $\mathrm{F}_{1}\%$ & Time$\%$ &  Acc$\%$ & Time$\%$ & Acc$\%$ & Time$\%$ & Acc$\%$ & Time$\%$ \\
\cmidrule(lr){1-13}
BERT-base & 92.1 & 100 & 90.1 & 100 & 87.8 & 100 & 91.3 & 100 & 84.4/85.0 & 100 & 71.1 & 100 \\
\cmidrule(lr){1-13}
DeeBERT & 89.7 & 44.8 & 89.8 & 72.2 & 85.5 & 45.7 & 87.0 & 50.1 & 83.3/83.6 & 72.7 & 68.2 & 94.2 \\
DeeBERT+SD & 89.7 & 46.2 & 89.5 & 78.0 & 85.0 & 49.3 & 87.1 & 49.0 & 83.3/83.7 & 73.3 & 69.0 & 90.7 \\
RomeBERT & \cellcolor{cyan}\bf90.5 & \cellcolor{cyan}\bf26.8 & \cellcolor{cyan}\bf90.2 & \cellcolor{cyan}\bf66.7 & \cellcolor{cyan}\bf85.8 & \cellcolor{cyan}\bf42.9 & \cellcolor{cyan}\bf89.2 & \cellcolor{cyan}\bf34.6 & \cellcolor{cyan}\bf83.3/84.0 & \cellcolor{cyan}\bf52.7 & \cellcolor{cyan}\bf70.8 & \cellcolor{cyan}\bf88.9  \\
\cmidrule(lr){1-13}
DeeBERT & 87.4 & 37.3 & 89.3 & 64.9 & 83.0 & 37.7 & 85.0 & 42.8 & 78.8/79.3 & 58.7 & 67.9 & 90.0 \\
DeeBERT+SD & 87.7 & 37.2 & 88.8 & 70.1 & 82.8 & 40.8 & 85.3 & 43.8 & 78.8/79.6 & 58.2 & 68.6 & 87.9 \\
RomeBERT & \cellcolor{cyan}\bf88.5 & \cellcolor{cyan}\bf18.6 & \cellcolor{cyan}\bf89.3 & \cellcolor{cyan}\bf59.1 & \cellcolor{cyan}\bf83.8 & \cellcolor{cyan}\bf26.7 & \cellcolor{cyan}\bf86.3 & \cellcolor{cyan}\bf24.7 & \cellcolor{cyan}\bf82.1/82.8 & \cellcolor{cyan}\bf38.6 & \cellcolor{cyan}\bf71.1 & \cellcolor{cyan}\bf69.4 \\
\cmidrule(lr){1-13}
DeeBERT  & 85.9 & 31.1 & 86.0 & 39.5 & 70.4 & 20.5 & 81.5 & 34.4 & 75.9/76.1 & 53.4 & 66.4 & 78.7 \\
DeeBERT+SD & 85.5 & 31.5 & 85.7 & 39.5 & 69.9 & 21.6 & 81.4 & 34.7 & 75.7/76.6 & 51.8 & 67.5 & 71.9 \\
RomeBERT & \cellcolor{cyan}\bf86.6 & \cellcolor{cyan}\bf14.2 & \cellcolor{cyan}\bf86.6 & \cellcolor{cyan}\bf32.7 & \cellcolor{cyan}\bf81.6 & \cellcolor{cyan}\bf16.5 & \cellcolor{cyan}\bf83.8 & \cellcolor{cyan}\bf20.1 & \cellcolor{cyan}\bf78.1/78.2 & \cellcolor{cyan}\bf21.1 & \cellcolor{cyan}\bf67.5 & \cellcolor{cyan}\bf43.9 \\
\bottomrule
\end{tabular}
}
\caption{Performance comparison on the \textsc{validation} splits of the GLUE benchmark between DeeBERT and RomeBERT.} 
\label{tab:valid}
\end{table*}

\subsection{Self-Distillation (SD) for RomeBERT}
DeeBERT freezes the BERT backbone when training the multi-exit classifiers. This strategy stabilizes the training process of multi-exit Transformer at the price of limiting the expressive power of BERT for early exits. To utilize the expressive power of BERT backbone, we propose to train intermediate classifiers with the final classifier jointly in one stage. However, naively unfreezing the BERT backbone may lead to conflicts between different exits, which will result in performance degradation for all exits. To solve the problem, we propose an additional consistency regularization for all exits beyond the original final-layer cross-entropy supervision from $\mathcal{L}_{final}=\mathcal{L}_{ce}(y, f_k(x;\theta_k))$. Specifically, we enforce the prediction of early exits to mimic the soft prediction of the final exit. The SD strategy can regularize the prediction consistency between all exits and transfer knowledge from the late exit to early exits, thus we can stabilize the training process and enable significant performance improvements for early exits. The SD strategy is shown in the left part of Figure~\ref{fig:model}, all the intermediate exits are supervised by both the soft label (i.e., logit) of the final classifier and the ground-truth label. Specifically, the SD loss ${\cal L}_{sd}$ contains two components: multi-exit cross-entropy loss ${\cal L}_{multi}$ and Kullback-Leibler divergence (KLD) loss ${\cal L}_{kld}$, it can be written as:
\begin{align}
	{\cal L}_{sd} & = {\cal L}_{multi} + {\cal L}_{kld} \nonumber \\
	  & = \sum_{i=1}^{k-1} [ (1-\gamma) \cdot {\cal L}_{ce} (y, f_i(x;\theta_i))   \nonumber \\ 
	  & + \gamma \cdot {\cal L}_{kld}^{(i)} (f_k(x;\theta_k) \parallel f_i(x;\theta_i), T) ], \nonumber
\end{align}
where $\gamma$ is a coefficient to balance the two loss terms and $T$ is the temperature of the KLD loss.

\begin{figure*}
\begin{center}
\includegraphics[width=0.99\linewidth,trim={0cm 0cm 0cm 0cm}]{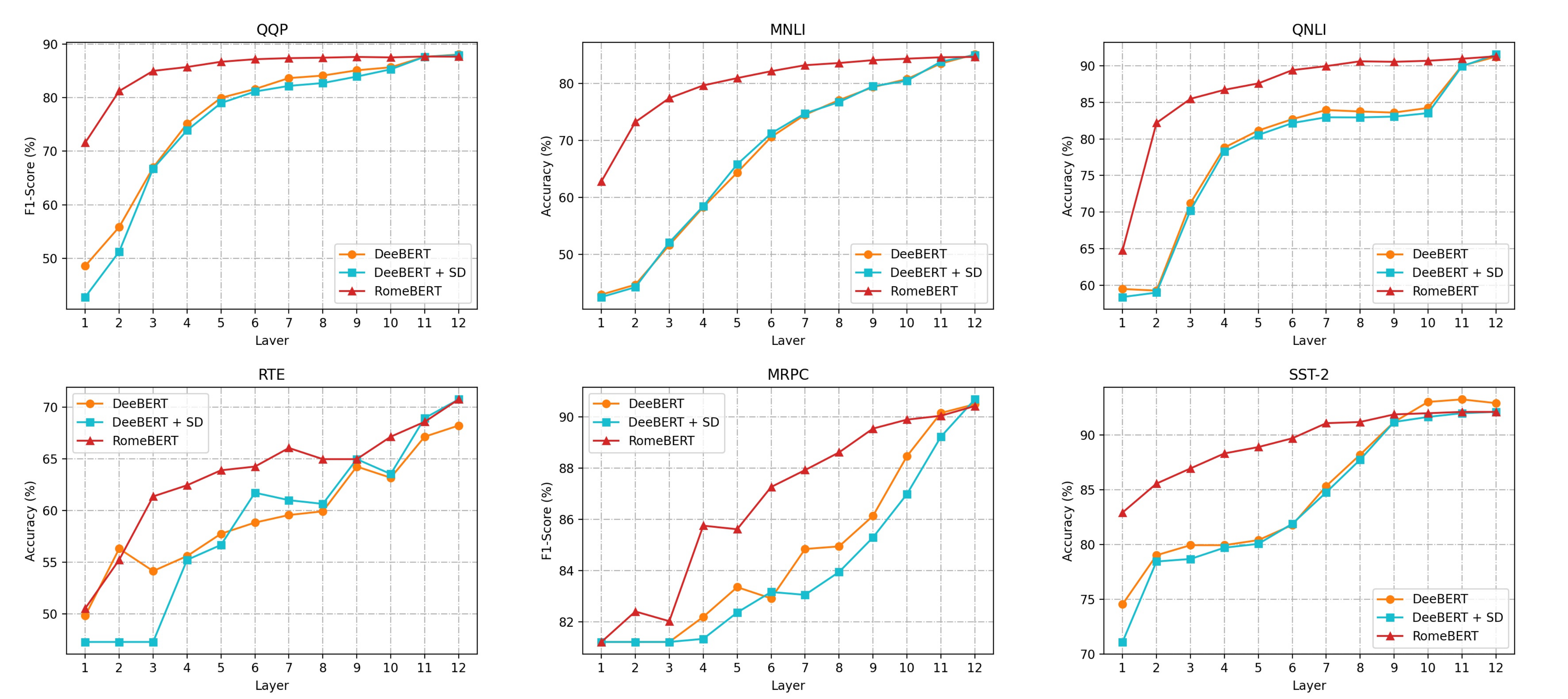}
\end{center}
\caption{Performance comparison on the \textsc{validation} splits of the six GLUE datasets among DeeBERT, DeeBERT+SD and RomeBERT.}
\label{fig:curve} 
\vspace{-0.3cm}
\end{figure*}

Suppose there are $\mathcal{C}$ classes in a classification task, we can denote the $i$-th layer Kullback-Leibler divergence loss ${\cal L}_{kld}^{(i)}$ as:
\begin{align}
	 {\cal L}_{kld}^{(i)} = - \sum_{c \in \mathcal{C}} p_k(c\mid x;\theta_k, T) \log\frac{p_i(c\mid x;\theta_i, T)}{p_k(c\mid x;\theta_k, T)} ,\nonumber
\end{align}
where $c$ is a class in $\mathcal{C}$, $T$ is a temperature value, and $p_i(c\mid x;\theta_i, T)$ can be computed as:
\begin{align}
	 p_i(c\mid x;\theta_i, T) = \frac{\exp(f_i(c \mid x;\theta_i)/T)}{\sum_j \exp(f_i(j \mid x;\theta_i)/T)}
	 . \nonumber
\end{align}
It is worth noting that FastBERT also adopts self-distillation to transfer knowledge from late to earlier exits. However, FastBERT follows the two-stage training paradigm of DeeBERT, where the BERT backbone is frozen during self-distillation in the second stage. In this paper, we implement DeeBERT plus self-distillation (DeeBERT+SD\footnote{Code is available at \url{https://github.com/romebert/DeeBERT-SD}.}) to reproduce the performance of FastBERT.

\subsection{\mbox{Gradient Regularization for RomeBERT}}
Self-distillation can stabilize training by enforcing consistent regularization over all exits. However, we still observe performance degradation on some GLUE tasks, i.e., RTE, QNLI and MRPC. We hypothesize that self-distillation can ease the conflicts among the training objectives of different exits but the new ${\cal L}_{sd}$ may still suffer from gradient conflicts with the final-exit training objective in these tasks.

Built upon the hypothesis, Gradient Regularization (GR) is proposed to further facilitate self-distillation in RomeBERT.
As shown in Figure~\ref{fig:model}, gradient conflict will arise when the angle between gradients computed by the final-exit loss and the combination of multi-exit loss and KLD loss is larger than $90^{\circ}$.
We denote the gradients of the final-exit loss and the self-distillation loss as $\mathbf{g}_f = \frac{\partial \mathcal{L}_{final}}{\partial \theta}$ and $\mathbf{g}_s = \frac{\partial \mathcal{L}_{sd}}{\partial \theta } =\frac{\partial (\mathcal{L}_{multi} +\mathcal{L}_{kld})}{\partial \theta }$, respectively.
To alleviate gradient conflict, GR will project the gradient of $\mathbf{g}_f$ to the normal direction of $\mathbf{g}_s$ when the angle between $\mathbf{g}_f$ and $\mathbf{g}_s$ is larger than $90^{\circ}$, i.e., $\mathbf{g}_f \cdot \mathbf{g}_s < 0$.
The projection can be formulated as:
\vspace{-0.2cm}
\begin{align}
	\mathrm{Proj}(\mathbf{g}_f) = \mathbf{g}_f - \frac{\mathbf{g}_f \cdot \mathbf{g}_s}{\left \| \mathbf{g}_s \right \|^2} \cdot \mathbf{g}_s. \nonumber
\end{align}
$\mathrm{Proj}(\mathbf{g}_f)$ decreases the gradient from the final-layer exit loss while not hurting SD, which can help to produce better intermediate exits. 
The modified gradient $\mathbf{g}^\star = \mathrm{Proj}(\mathbf{g}_f) + \mathbf{g}_s$ will then be used to update the model parameters.
In the other case (when the angle between $\mathbf{g}_f$ and $\mathbf{g}_s$ is less than $90^{\circ}$, i.e., $\mathbf{g}_f \cdot \mathbf{g}_s > 0$), the original gradients $\mathbf{g}_f$ and $\mathbf{g}_s$ will be kept as Case 2 shown in Figure~\ref{fig:model}.

\begin{figure}[htbp!]
\begin{center}
\includegraphics[width=1.02\linewidth,trim={0cm 0cm 0cm 0cm}]{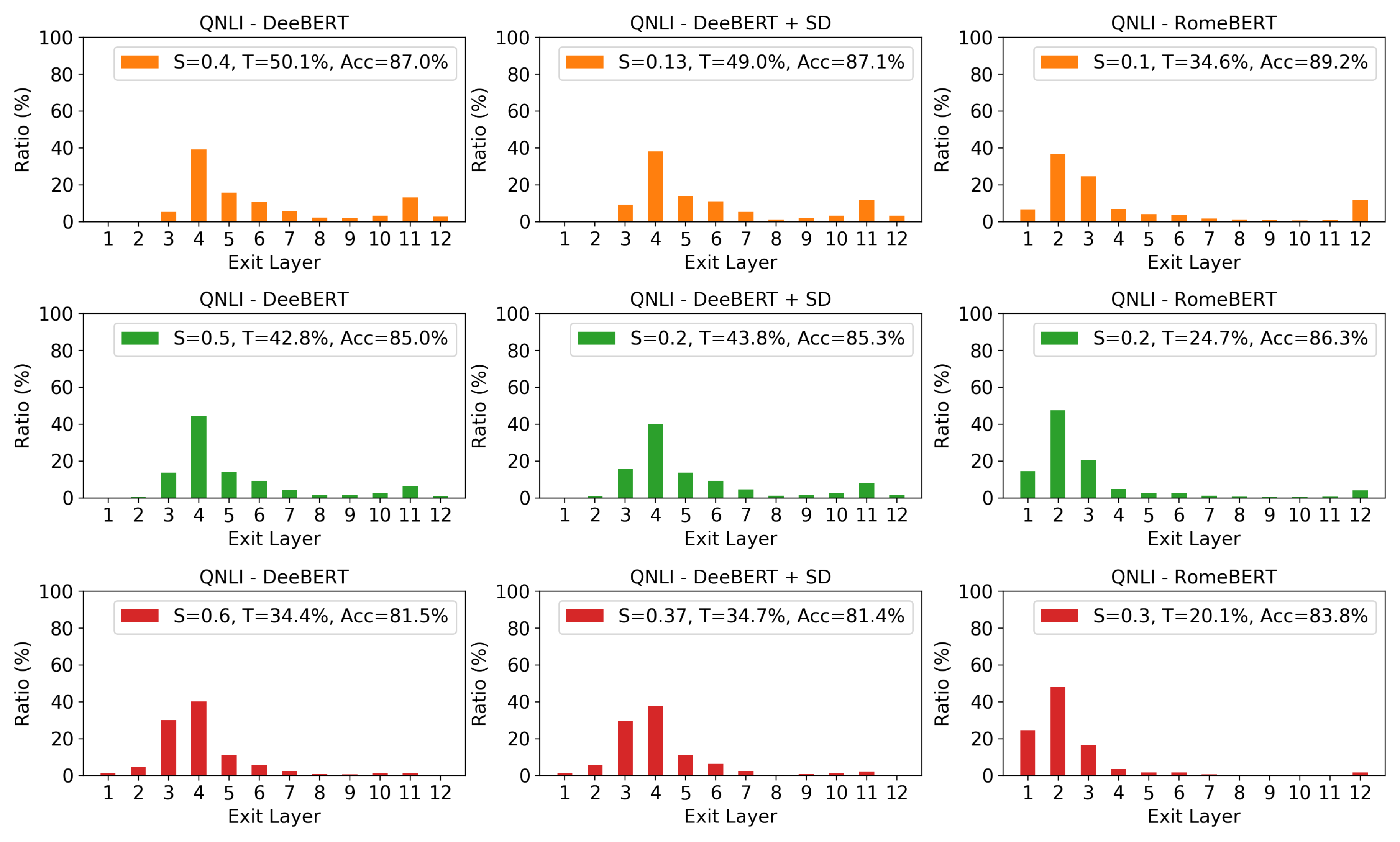}
\end{center}
\caption{Exit-layer distributions of DeeBERT, DeeBERT+SD and RomeBERT on QNLI validation split.}
\vspace{-0.2cm}
\label{fig:qnli}
\end{figure}

\begin{figure}[htbp!]
\begin{center}
\includegraphics[width=1.02\linewidth,trim={0cm 0cm 0cm 0cm}]{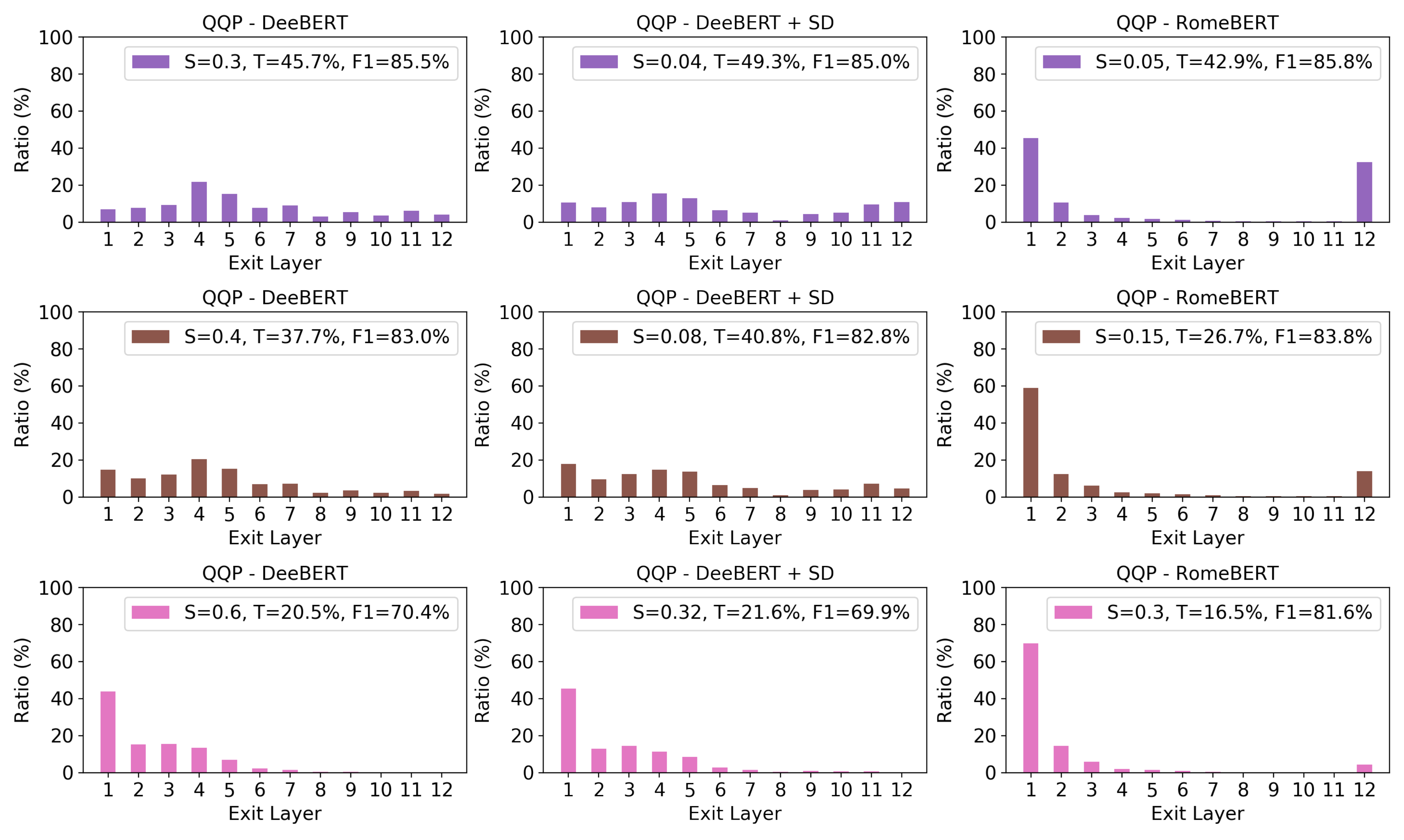}
\end{center}
\caption{Exit-layer distributions of DeeBERT, DeeBERT+SD and RomeBERT on QQP validation split.}
\label{fig:qqp}
\vspace{-0.15cm}
\end{figure}


\begin{figure}[htbp!]
\begin{center}
\includegraphics[width=1.02\linewidth,trim={0cm 0cm 0cm 0cm}]{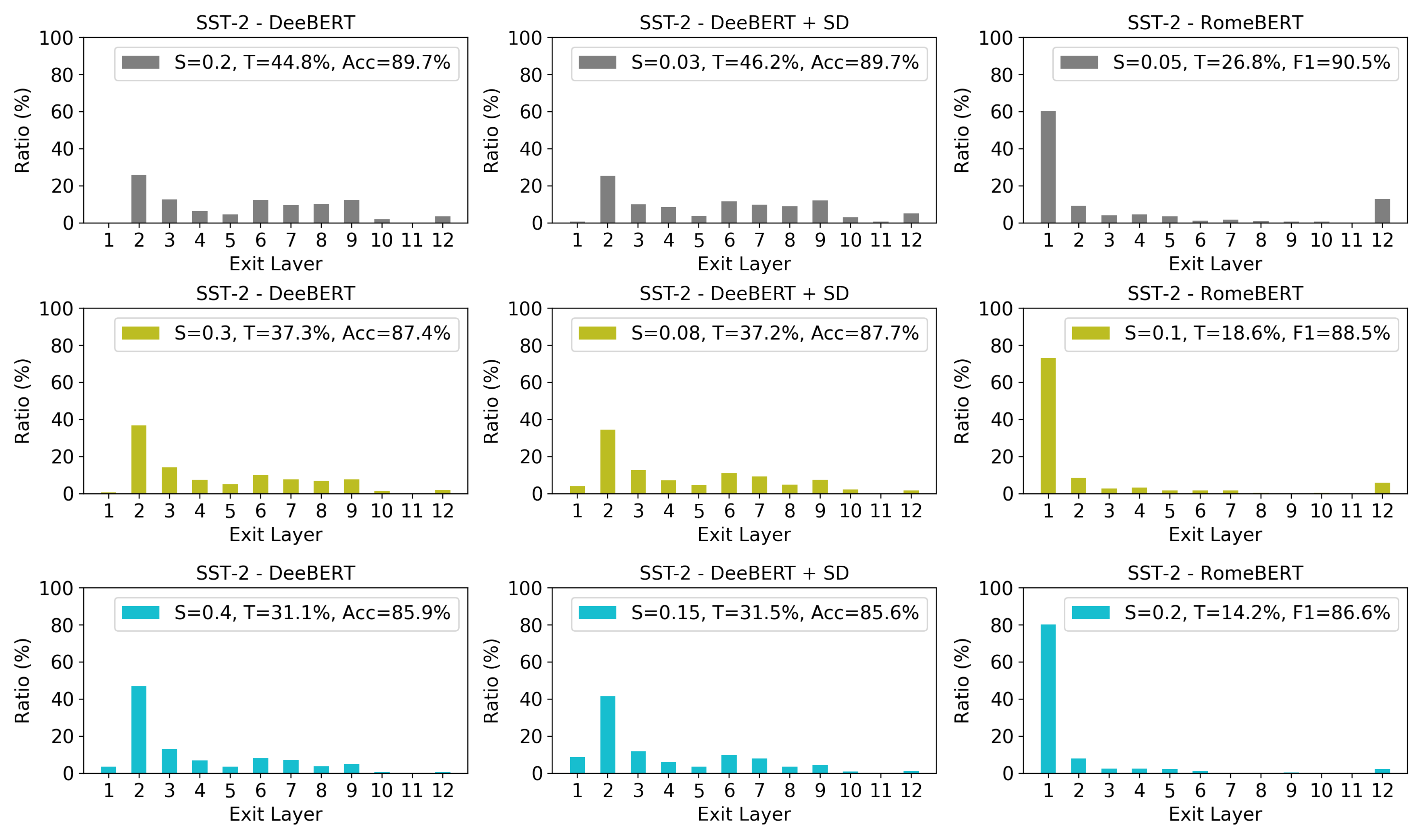}
\end{center}
\caption{Exit-layer distributions of DeeBERT, DeeBERT+SD and RomeBERT on SST-2 validation split.}
\label{fig:sst-2}
\vspace{-0.3cm}
\end{figure}



\section{Experiments}
\subsection{Datasets and Experimental Setup}
\label{sec:datasets}
We train all approaches from the BERT-base pre-trained model~\cite{devlin2018bert}, and conduct experiments on the same GLUE benchmark \citep{wang2018glue} that is used by DeeBERT: QQP, MNLI, MRPC, RTE, SST-2, and QNLI. In our experiments, we maintain the same hyperparameters as DeeBERT. Specifically, we set coefficient $\gamma$ to 0.9, self-distillation temperature $T$ to 3.0 and keep them unchanged for all experiments. 

Our experiments focus on two aspects of early-exit performance improvements: 1) Group different models with similar performances by changing their adaptive inference entropy values, then compare the expected running time for these models. We follow this setup to compare RomeBERT with DeeBERT and DeeBERT+SD on both validation and test splits. We also visualize the exit-layer distribution for each approach to show the advantage of RomeBERT.
2) Fix the exit layer and calculate the exit performance for all layers. Based on this setting, we compare RomeBERT with DeeBERT and DeeBERT+SD on the validation split, then conduct layerwise ablation study for the two key components of RomeBERT. 

To make better comparison, we summarize the core properties of each approach. In terms of training stage, DeeBERT and DeeBERT+SD follow the two-stage training strategy while RomeBERT conducts joint training in only one stage. DeeBERT and DeeBERT+SD both freeze the BERT backbone during the second training stage. In terms of self-distillation, DeeBERT+SD conducts self-distillation in the second stage while RomeBERT performs self-distillation and gradient regularization together within one stage.

\begin{figure*}
\begin{center}
\includegraphics[width=0.99\linewidth,trim={0cm 0cm 0cm 0cm}]{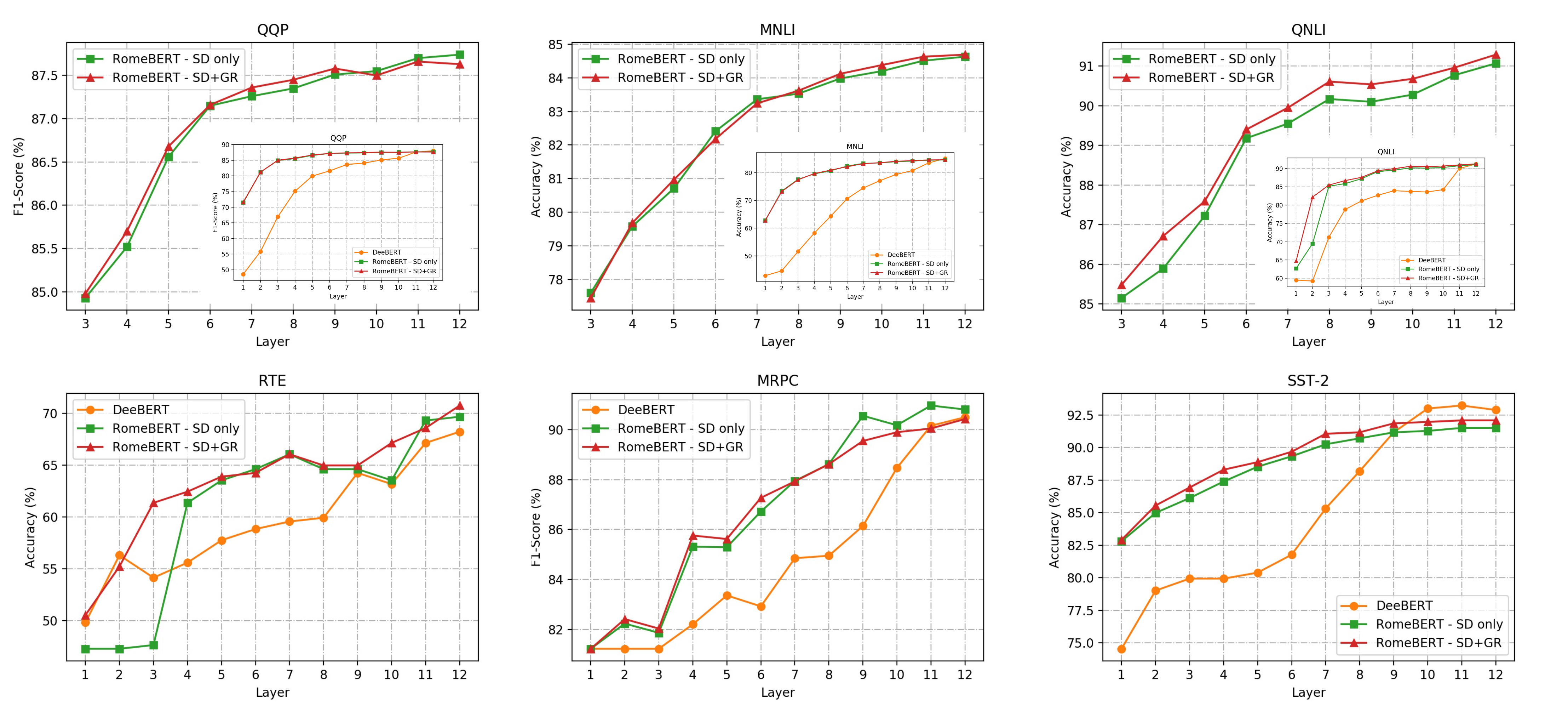}
\end{center}
\caption{Performance comparison on the \textsc{validation} splits of the six GLUE datasets between DeeBERT and RomeBERT. Ablation study on the two key components of RomeBERT is also included. For three datasets (QQP, MNLI, and QNLI), we enlarge details to better compare the performance of SD only with SD+GR.}
\vspace{-0.3cm}
\label{fig:ablation}
\end{figure*}

\subsection{Performance Comparison and Discussion}
\label{sec:compare}
The performance on test splits of GLUE is shown in Table \ref{tab:main}. Here we regard the expected running time of BERT-base as 100\%, and calculate the relative expected running time percentage for DeeBERT, DeeBERT+SD and RomeBERT in adaptive inference mode. If an input example early exits at layer $i$, its running time is $i/k$ of the situation that exits at the final layer $k$. In Table \ref{tab:main}, we group results from different approaches with similar performances, then compare their expected running time. From the table, we can observe that in most datasets and situations, RomeBERT can achieve better performance than DeeBERT/DeeBERT+SD while requiring less running time. 

For small datasets such as RTE, RomeBERT can achieve 69.5\% accuracy, which is 0.1\% better than the BERT-base baseline, and with 10.6\% time saving. In contrast, DeeBERT needs 5.4\% more time with 1.6\% worse performance, DeeBERT+SD requires 1.2\% more time with 1.9\% performance drop. In addition, RomeBERT gets 66.9\% accuracy with a 54.8\% time saving, while DeeBERT needs 33.8\% more time but the performance is 0.8\% worse, DeeBERT+SD requires 25.8\% more time but with 0.6\% performance drop. 

For middle-sized datasets such as QNLI, RomeBERT can achieve 88.7\% accuracy which is 2.1\% below BERT-base but with only 34.1\% running time. In contrast, DeeBERT obtains 87.1\% accuracy with 15.2\% more time, DeeBERT+SD gets only 86.6\% accuracy at the cost of 13.8\% more time.
If we constrain the expected running time to 34.1\%, DeeBERT can only achieve 82.4\% accuracy compared with RomeBERT's 88.7\%, and DeeBERT+SD can only obtain 81.7\% accuracy.

For large-sized datasets such as QQP, our RomeBERT obtains 70.8\% $\mathrm{F}_{1}$-score which is only 0.5\% less than BERT-base with a huge 65.2\% time saving. Even when the time cost is further reduced to 15.2\%, RomeBERT only drops 3.6\% $\mathrm{F}_{1}$-score than BERT-base, but DeeBERT needs 20.1\% more time to achieve the same performance. If we change the performance standard to 69.7\%, DeeBERT needs 42.6\% time to achieve this goal while RomeBERT only requires 22.8\%. Besides, in Table~\ref{tab:valid}, we also show the performances of different methods on validation splits of GLUE benchmark for reference. 

We illustrate the exit-layer distributions on QNLI, QQP and SST-2 in Figure~\ref{fig:qnli}, Figure~\ref{fig:qqp} and Figure~\ref{fig:sst-2}, respectively. From these figures, we can see that the exit-layer distribution of RomeBERT is more prone to earlier exits than DeeBERT and DeeBERT+SD. For QNLI, the exit layer distribution of DeeBERT/DeeBERT+SD is centered at the 4th layer, while RomeBERT is centered at the 2nd layer. For QQP, the exit layers of DeeBERT/DeeBERT+SD are more uniformly distributed. In contrast, the 1st layer gets the most frequent exits for RomeBERT.
For SST-2, the most frequent exit layer for DeeBERT/DeeBERT+SD is the 2nd layer, but the ratio is at most 40\%, while the 1st layer again gets the most frequent exits for RomeBERT and the ratio is at least 60\%. The observation proves that RomeBERT can really improve the performance and maintain a robust training for early exits.
Moreover, Figure~\ref{fig:curve} shows every-layer performance comparison when the exit layer is fixed for all inputs. We can see that the curves of RomeBERT are above those of DeeBERT and DeeBERT+SD in almost all situations. The performance gaps between RomeBERT and DeeBERT/DeeBERT+SD are especially large for early layers and on large-sized datasets. At the first three layers, for instance, RomeBERT increases nearly 20\% performance over DeeBERT/DeeBERT+SD on both QQP and MNLI. We can also conclude that DeeBERT+SD does not beat DeeBERT in terms of layerwise performance. This implies that freezing the BERT backbone during self-distillation may limit the expressive power of early exits.

\subsection{Ablation Study on RomeBERT}
\label{sec:ablation}
In Figure~\ref{fig:ablation}, we also conduct ablation study on GR and SD. Since GR is built upon SD, we directly compare the performances of RomeBERT with or without GR. Due to the large performance gap in early layers of RomeBERT over DeeBERT, we rescale the curves to better evaluate the function of GR. From the figure, we can conclude that GR is generally helpful to improve the performance upon SD. The degree of improvement depends on the level of gradient conflict. Specifically, GR exhibits its superiority on RTE and QNLI since almost all layers get improved accuracy. In addition, GR improves the performance of most layers on QQP and SST-2, while it is partially effective for certain layers on MRPC and MNLI.

\section{Conclusion}
In this paper, we propose RomeBERT for robust training of multi-exit BERT. We validate the effectiveness and efficiency of RomeBERT on six GLUE classification tasks. Compared with the DeeBERT and FastBERT approaches, RomeBERT, which is driven by self-distillation and gradient regularization to facilitate the early exits, achieves better speed-performance tradeoff on all tasks.

\bibliographystyle{acl_natbib}
\bibliography{emnlp2020}

\end{document}